\documentclass[twoside, journal]{IEEEtran}
\usepackage{times}
\usepackage{epsfig}
\usepackage{graphicx}
\usepackage{amsmath}
\usepackage{amssymb}
\usepackage{multirow}
\usepackage{color}
\usepackage{amssymb}
\usepackage{cite}
\newcommand{\RomanNumeralCaps}[1]
{\MakeUppercase{\romannumeral #1}}
\newcommand{\etal}{\textit{et al}. }
\newcommand{\ie}{\textit{i}.\textit{e}., }
\newcommand{\eg}{\textit{e}.\textit{g}., }
\begin{document}

\title{Relation Regularized Scene Graph Generation}

\author{Yuyu~Guo,
        Lianli~Gao,
        Jingkuan~Song,
        Peng~Wang,
        Nicu~Sebe,
        Heng~Tao~Shen~\IEEEmembership{Fellow,~ACM}, 
        and Xuelong~Li~\IEEEmembership{Fellow,~IEEE}% <-this % stops a space

% <-this % stops a space
%\thanks{Manuscript received July 16, 2019; revised October 17, 2020 and May, 13, 2020; accepted January 2, 2021. This work was supported in part by the Fundamental Research Funds for the Central Universities under Grant ZYGX2014J063 and Grant ZYGX2014Z007; and in part by the National Natural Science Foundation of China under Grant 61502080, Grant 61632007 and Grant 61602049. This article was recommended by Associate Editor \textcolor{red}{***}. (\textit{Corresponding author: \textcolor{red}{***}.)}}
\thanks{Yuyu~Guo, Lianli~Gao, Jingkuan~Song and Heng~Tao~Shen are with the Future Media Center and School of Computer Science and Engineering, University of Electronic Science and Technology of China, Chengdu 611731, China  (e-mail: yuyuguo1994@gmail.com, lianli.gao@uestc.edu.cn, jingkuan.song@gmail.com, shenhengtao@hotmail.com).	Corresponding Author: Lianli Gao.}% <-this % stops a space
\thanks{Peng~Wang is with the School of Computing and Information Technology, The University of Wollongong, NSW 2170, Australia (e-mail: pengw@uow.edu.au).}% <-this % stops a space
\thanks{Nicu~Sebe is with the Department of Information Engineering and Computer
	Science, University of Trento, 38123 Trento, Italy (e-mail: niculae.sebe@unitn.it).}% <-this % stops a space
\thanks{Xuelong~Li is with School of Computer Science and Center for OPTical IMagery Analysis and Learning (OPTIMAL), Northwestern Polytechnical University, Xi'an 710072, P.R. China (e-mail: xuelong\_li@nwpu.edu.cn). }% <-this % stops a space
%\thanks{Color versions of one or more figures in this article are available https://doi.org/\textcolor{red}{****}}% <-this % stops a space
%\thanks{Digital Object Identifier \textcolor{red}{****}}% <-this % stops a space
% \thanks{Manuscript received April 19, 2005; revised August 26, 2015.}
}

% and the end of the author line. \ie if you had this:

\markboth{IEEE TRANSACTIONS ON CYBERNETICS}%
{Guo \MakeLowercase{\textit{et al.}}: Relation Regularized Scene Graph Generation}
%
%{Shell \MakeLowercase{\textit{et al.}}: Bare Demo of IEEEtran.cls for %IEEE Journals}



\maketitle

\begin{abstract}
Scene graph generation (SGG) is built on top of detected objects to predict object pairwise visual relations for describing the image content abstraction. Existing works have revealed that if the links between objects are given as prior knowledge, the performance of SGG is significantly improved. Inspired by this observation, in this paper, we propose a Relation Regularized Network (R2-Net), which can predict whether there is a relationship between two objects and encode this relation into object feature refinement and better SGG. Specifically, we first construct an affinity matrix among detected objects to represent the probability of a relationship between two objects. Graph Convolution Networks (GCNs) over this relation affinity matrix are then used as object encoders, producing relation-regularized representations of objects. With these relation-regularized features, our R2-Net can effectively refine object labels and generate scene graphs. Extensive experiments are conducted on the Visual Genome dataset for three SGG tasks (\ie PREDCLS, SGSLS, and SGDET), demonstrating the effectiveness of our proposed method. Ablation studies also verify the key roles of our proposed components in performance improvement.
\end{abstract}


%\ifCLASSOPTIONpeerreview
%%\begin{center} \bfseries EDICS Category: 3-BBND \end{center}
%\fi
%\IEEEpeerreviewmaketitle

\section{Introduction}
\label{sec:int}
\begin{figure}
	\centering
	\includegraphics[width=0.7\linewidth]{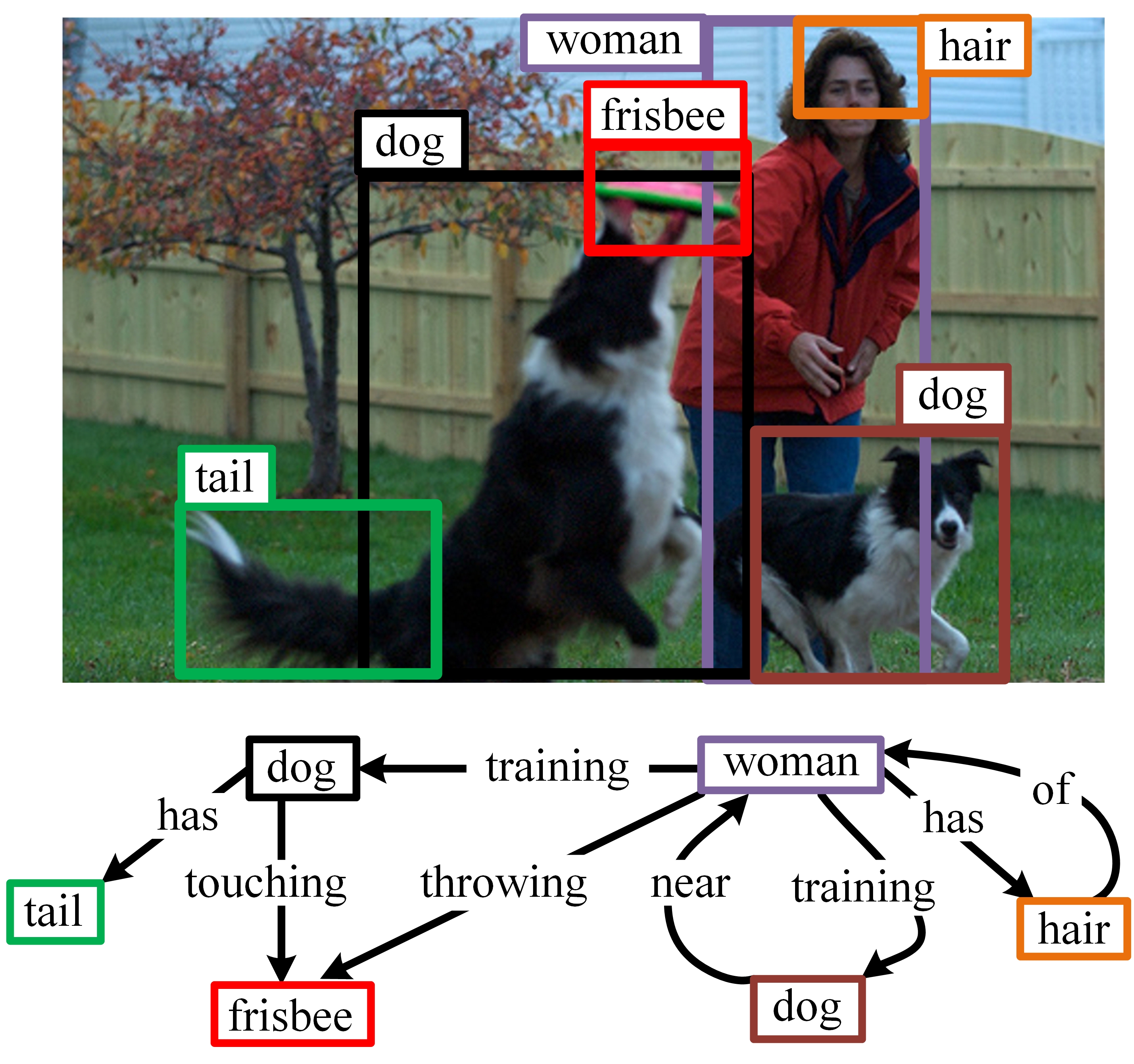}
	\caption{An ideal scene graph generation method takes an image as input to generate a precise graph for describing the image content abstraction. A scene graph consists of nodes (dog, tail and so on) and edges (has, touching and so on). It can also be represented by a set of triples (\textit{dog-has-tail}, \textit{dog-touching-frisbee} and so on).}
	\label{Fig.demo}
\end{figure}

% why scene graph
\IEEEPARstart{I}{n practice} providing only object labels and detecting object bounding boxes~\cite{classify:googlenet,classify:VGG,classify:resnet,obj_det:yolo,obj_det:faster_rcnn} may not produce satisfactory semantic information for downstream tasks, such as visual content retrieval~\cite{img_ret:scenegraph,img_ret:cm_att,img_ret:ssv,img_ret:eac}, visual question answering~\cite{vqa:man, vqa:zpp_efa} and visual captioning~\cite{img_cap:bottmup, vid_cap:hat}. For instance, in Fig.~\ref{Fig.demo}, generated object labels and bounding boxes (\eg dog, woman, frisbee, and hair) cannot provide answers to the following question: \textit{What are the two dogs and the woman doing?} As a result, scene graph generation has been proposed and studied by Krishna~\etal~\cite{scenegraph:visual_genome}, who also collected a dataset consisting of images with objects and object relations to evaluate the quality of the generated scene graph. In the example in  Fig.~\ref{Fig.demo}, the bottom part represents the scene graph, precisely and briefly describing the semantic content of the top image. With such a scene graph, we can provide an answer to the aforementioned question. 
% how to conduct scene graph
As shown in Fig.~\ref{Fig.demo}, a scene graph consists of nodes (objects) and edges (relationships between objects).
Given an image, scene graph generation is built on top of detected objects to predict object pairwise visual relations for describing the image content abstraction~\cite{scenegraph:VRD_LP,scenegraph:neural_motifs,scenegraph:IMP}. It is challenging to effectively and accurately generate a scene graph and this has been recently the subject of intensified research~\cite{scenegraph:VRD_LP,scenegraph:VTEN,scenegraph:IMP,scenegraph:neural_motifs,scenegraph:factnet}. Some studies have focused on exploiting linguistic priors~\cite{scenegraph:VRD_LP}, visual embedding~\cite{scenegraph:VTEN}, feature interactions~\cite{scenegraph:zoomnet}, external information (\eg region captions)~\cite{scenegraph:denscap} or spatial information~\cite{scenegraph:factnet} to boost the performance. Some other studies have tried to extract certain contextual information from objects to enhance the performance~\cite{scenegraph:IMP,scenegraph:neural_motifs,scenegraph:graphrcnn} by iterative message passing, neural motifs, or graphs to avoid detecting and recognizing individual objects in isolation. 

From previous works, we can see that the contextual information is important to scene graph generation.
To truly produce accurate scene graphs, it is crucial to devise a model that fully and automatically exploits global contextual information and relational contextual information. The global contextual information encodes the visual information from all objects and backgrounds/environments. The relational contextual information encodes the graph information among objects. 
Moreover, the previous experimental results of the two scene graph generation tasks (\ie \textit{PREDCLS} and \textit{SGCLS})~\cite{scenegraph:IMP,scenegraph:neural_motifs,scenegraph:factnet} have proved that the quality of the detected object labels significantly influences the performance of the scene graph generation. In other words, improving the quality of object label detection could directly lead to better scene graphs.

Therefore, in this paper, we address the problem of scene graph generation, where we aim to take advantage of the global context and relational information to produce a relation regularized scene graph from an image. Our contributions are three folds: 1) We propose a novel relation regularized network, namely R2-Net, which can predict whether there is a relationship between two objects and use this relation as a regularizer to learn relation-embedded features. Therefore, the R2-Net effectively and progressively encode region features with the global context and relational information to refine object label prediction and scene graph generation;  2) We propose a stacked LSTM-GCN encoder to extract the comprehensive features of objects. Specifically, we stack GCNs on top of Bi-LSTMs as an object feature encoder to combine the relation-embedded features and the global features. 3)  We verify the effectiveness of our method across three tasks (\ie PREDCLS, SGSLS, and SGDET) on the Visual Genome dataset. Our ablation study also demonstrates the key roles of our proposed components in performance improvement.

\section{Related Work}
In this section, we describe three categories of related works, including object detection, scene graph generation and graph convolutional networks.

\textbf{Object Detection.} Object detection is one of the most fundamental areas in the field of computer vision. Owing to the evolution of Convolutional Neural Networks (CNNs)~\cite{network:CNNs},  many effective CNN-based methods~\cite{obj_det:faster_rcnn,obj_det:yolo,obj_det:rcnn,obj_det:fastrcnn,obj_det:ssd,obj_det:maskrcnn} have been proposed to deal with this task. At first, Girshick~\etal~\cite{obj_det:rcnn} directly extracted the deep features of warped region proposals with a deep CNN and classified these deep region features with SVMs~\cite{ml:svm}. However, performing a whole CNN forward pass for each proposal is time-consuming. Fast R-CNN~\cite{obj_det:fastrcnn} extracted the feature map from an entire image and mapped region proposals to regions of interest (RoIs) in the feature map. Then, the RoIPooling layer resized the RoI feature maps to the same size for classification and regression. By sharing the same CNN for region proposals, Fast R-CNN saved a lot of computing time. In order to achieve real-time detection, Faster R-CNN~\cite{obj_det:faster_rcnn} replaced the time-consuming Selective Search method with a region proposal network to search for the class-agnostic foreground objects. Previous methods with the RoI pooling layer may cause coarse quantization on feature maps. To alleviate this problem, He~\etal~\cite{obj_det:maskrcnn} proposed the RoIAlign layer, which preserves precise spatial mappings. Different from the above works, this paper focuses on a more complex problem: scene graph generation. To solve this problem, we need not only to detect objects in the image but also to extract contextual information to predict the relationships among objects.

\textbf{Scene Graph Generation.} Object detection cannot adequately represent rich semantic information in images, and therefore
more and more works~\cite{scenegraph:IMP,scenegraph:VTEN,scenegraph:visual_genome,scenegraph:VRD_LP,scenegraph:factnet,scenegraph:denscap,scenegraph:drnet,scenegraph:oneshot} pay attention to scene graph generation (or visual relationship detection) for exploring rich semantic information in images. Since the relationships between objects largely depend on human prior knowledge, 
Lu~\etal~\cite{scenegraph:VRD_LP} integrated a visual module and a language module for adequately employing human prior knowledge.
Inspired by translating embeddings~\cite{mrd:transe} for modeling multi-relational data, Zhang~\etal~\cite{scenegraph:VTEN} proposed the Visual Translation Embedding (VTE) network. In VTE, entities are embedded in a semantic space, and relationships are modeled as a vector translation: subject - object $ \approx $ relation. 
Due to the effectiveness of end-to-end convolutional neural networks, Newell~\etal~\cite{scenegraph:PGAE} proposed end-to-end convolutional networks that mapped pixels to graphs directly with associative embedding.

All the above works use the local detectors and independently predict relationships between entities. 
As mentioned in~\cite{scenegraph:IMP}, ignoring the surrounding context may lead to ambiguity of model prediction. Therefore, for capturing the surrounding context in images, the authors inferred scene graphs by iteratively refining model predictions with recurrent neural networks. In addition, Zeller~\etal~\cite{scenegraph:neural_motifs} explored regularly appearing substructures called \textit{motifs} in scene graphs. Inspired by this analysis, the authors proposed a strong baseline. The baseline determines the relationship between two objects through two steps: 1. Determining the labels of objects by Faster R-CNN~\cite{obj_det:faster_rcnn}; 2. Finding the most frequent relationship between the two objects' labels (ignoring the visual information) in the training set. Then the authors combined the baseline and LSTMs for extracting global context and outperformed the state-of-the-art methods.
By analyzing the above works, we find that if the links between objects are given as prior knowledge, the performance of scene graph generation would be significantly improved.
Inspired by this observation, we propose a Relation Regularized Network (R2-Net), which can predict whether there is a relationship between two objects, and encode this relation with GCNs to refine features and generate robust scene graphs.

\textbf{Graph Convolutional Networks (GCNs).} In order to extract features on graph-structured data, Kipf~\etal~\cite{networks:gcn_kipfs_semi} proposed Graph Convolutional Networks (GCNs) for semi-supervised node classification based on spectral graph convolutions. Given a graph, GCNs refine node features based on the adjacency matrix and encode graph structures by passing information between adjacent nodes. Due to the effectiveness of GCNs, several works~\cite{networks:gcn_rel,networks:gcn_srl,hoi:gcns} introduced GCNs into different fields. For instance, Marcheggiani~\etal~\cite{networks:gcn_srl} used GCNs for semantic role labeling (SRL). They stacked GCNs on LSTMs to capture different ranges of information. Schlichtkrull~\etal~\cite{networks:gcn_rel} proposed Relational Graph Convolutional Networks (R-GCNs) for link prediction and entity classification on knowledge graphs.
In addition to the graph structure data, GCNs were also applied to computer vision tasks. In order to predict human-object interactions (HOIs) in images, Qi~\etal~\cite{hoi:gcns} expressed HOI structures as graphs. The information between instances can be effectively captured by GCNs. Different from these works, we focus on scene graph generation. Since the task of scene graph generation does not give the affinity matrix in the test phase, we first construct the affinity matrix from the image. Next, GCNs are used to encode the instance features with the affinity graph. In this way, our model can generate robust scene graphs with the graph level features.

\begin{table*}[]
\begin{center}	
\caption{The Table of Main symbols.}
\begin{tabular}{|l|l|l|}
\hline
\multicolumn{1}{|c|}{Symbol Name}                    & \multicolumn{1}{c|}{Dimension} & \multicolumn{1}{c|}{Description}                                                                     \\ \hline
$D_l$                                                & $D_l \in \mathbb{Z}^{+}$       & Total number of object categories.                                                                   \\ \hline
$D_f$                                                & $D_f \in \mathbb{Z}^{+}$       & Dimension of features extracted from Faster R-CNN.                                                   \\ \hline
$D_h$                                                & $D_h \in \mathbb{Z}^{+}$       & Dimension of features extracted from Bi-LSTMs.                                                       \\ \hline
$D_o$                                                & $D_o \in \mathbb{Z}^{+}$       & Dimension of output features of the reltion regulrized encoder in the object lable refiner.          \\ \hline
$D_z$                                                & $D_z \in \mathbb{Z}^{+}$       & Dimension of output features of the reltion regulrized encoder in the object relationship generator. \\ \hline
$D_r$                                                & $D_r \in \mathbb{Z}^{+}$       & Total number of relationship predicate categories.                                                   \\ \hline
$\textbf{B} = \{b_1,...,b_N\}$                       & $b_i \in \mathbb{R}^{4}$       & Bounding boxes generated from Faster R-CNN.                                                          \\ \hline
$\textbf{L} = \{l_1,...,l_N\}$                       & $l_i \in \mathbb{R}^{D_l}$     & Label probabilities generated from Faster R-CNN.                                                     \\ \hline
$\textbf{F} = \{f_1,...,f_N\}$                       & $f_i \in \mathbb{R}^{D_f}$     & Object features extracted from Faster R-CNN.                                                         \\ \hline
$\textbf{U} = \{u_{1,1},...,u_{N,N}\}$               & $u_{i,j} \in \mathbb{R}^{D_f}$ & Union features extracted from Faster R-CNN.                                                          \\ \hline
$\textbf{H} = \{h_1, h_2, ..., h_N\}$                & $h_i \in \mathbb{R}^{D_h}$     & Global features extracted from Bi-LSTMs.                                                             \\ \hline
$\textbf{A}^e = \{a^e_{1,1},...,a^e_{N,N}\}$         & $a^e_{i,j} \in \mathbb{R} $    & Affinity matrix of the reltion regulrized encoder in the object label refiner.                       \\ \hline
$\textbf{O}'=\{o'_1,...,o'_N\}$                       & $o'_i \in \mathbb{R}^{D_o}$    & Output features of the relation regularized encoder in the object label refiner.                     \\ \hline
$\textbf{A}^r = \{a^r_{1,1},...,a^r_{N,N}\}$         & $a^r_{i,j} \in \mathbb{R} $    & Affinity matrix of the reltion regulrized encoder in the object relationship generator.              \\ \hline
$\textbf{Z}=\left\{ {{z_1}, \cdots ,{z_N}} \right\}$ & $z_i \in \mathbb{R}^{D_z}$     & Output features of the relaion regulzeried encoder in the object relaionship generator.              \\ \hline
$\textbf{R} = \{r_{1,1,1},...,r_{D_r,N,N}\}$         & $r_{m,i,j} \in \mathbb{R}$     & Probability that the $i$-th object and the $j$-th object belong to the relationship predicate $m$.   \\ \hline
\end{tabular}
\end{center}	
\label{tab.symb}
\end{table*}
\begin{figure*}
	\begin{center}
		\includegraphics[width=1.0\linewidth]{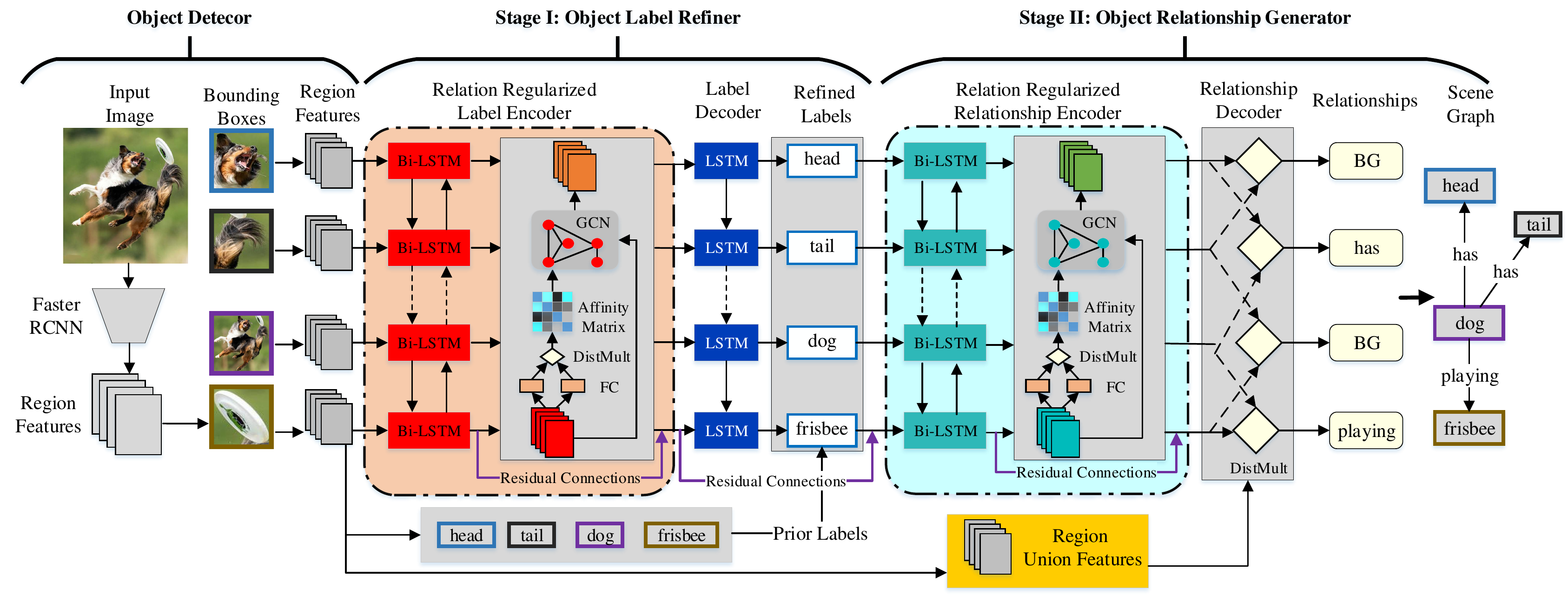}
	\end{center}
	\caption{The framework of Relation Regularized Network. After the object detector, our model can be divided into two stages: an object label refiner for refining the prior labels from the object detector and an object relationship generator for generating scene graphs.}
	\label{fig:framework}
\end{figure*}

\section{Relation Regularized Model}
\label{sec.method}
In this paper, we propose a relation regularized model for the scene graph generation. The overview of our proposed model is depicted in Fig.~\ref{fig:framework}. The framework consists of three components: 1) Object detection (\ie bounding box detection and prior object label detection); 2) Relation regularized label refiner; and 3) Relation regularized relationship generation. In the following subsections, we first introduce the definition of our problem and then describe the details of the model from inputs to outputs (Object Detector, Object Label Refiner, Object Relationship Generator, and Loss Functions). Since this section contains many symbols, we show the dimensions and descriptions of the main symbols we use in Tab.~\ref{tab.symb}.

\subsection{Problem Definition}
\label{sec.profor}
We define the image scene graph generation problem as follows: Given an image $\textbf{I}$, we want to generate a scene graph $\textbf{G}$ to describe its content abstraction. Following previous works~\cite{scenegraph:neural_motifs,img_ret:scenegraph}, we determine to progressively decompose the scene graph generation into a series of continued actions: bounding boxes detection $\textbf{B}$, bounding box label detection $\textbf{L}$ and label relation detection $\textbf{R}$. Therefore, $\textbf{G}$ is defined as $\textbf{G} = \left\{ {\textbf{B},\textbf{L},\textbf{R}} \right\}$. The probability of $\textbf{G}$ is decomposed by a multiplication rule: 
\begin{equation}
\begin{array}{l}
P(\textbf{G}|\textbf{I}) \\
=P(\textbf{B},\textbf{L},\textbf{R}|\textbf{I})\\
=P(\textbf{B}|\textbf{I})P(\textbf{L}|\textbf{B},\textbf{I})P(\textbf{R}|\textbf{L},\textbf{B},\textbf{I}).
\end{array}
\label{equ.dec_pro}
\end{equation}

\subsection{Object Detector}
\label{OBSTP1}
The object detector is utilized to detect instances in images. The inputs of the object detector are images, and the outputs are bounding boxes, categories, and features of instances. Faster R-CNN~\cite{obj_det:faster_rcnn} has achieved great success in image object detection, and it has been widely adopted to support image scene graph generation~\cite{scenegraph:neural_motifs,scenegraph:IMP,scenegraph:graphrcnn,scenegraph:VTEN}. In this paper, we adopt Faster R-CNN to generate a set of bounding boxes $\textbf{B} = \{b_1,...,b_N\}$, where $b_i \in \mathbb{R}^{4}$. $N$ is the total number of detected bounding boxes from the input image. $i$ is an index ranging from $1$ to $N$. With $N$ boxes, we can further obtain:
\begin{itemize}
	\item A set of label probabilities $\textbf{L} = \{l_1,...,l_N\}$, where $l_i \in \mathbb{R}^{D_l}$ and $D_l$ is the total number of labels in a dataset;
	\item A set of object feature vectors $\textbf{F} = \{f_1,...,f_N\}$, where $f_i \in \mathbb{R}^{D_f}$ and $D_f$ is the feature dimension;
	\item A set of feature vectors of union boxes $\textbf{U} = \{u_{1,1},...,u_{N,N}\}$, where $u_{i,j} \in \mathbb{R}^{D_f}$. Each union box is the smallest rectangle containing two bounding boxes.
\end{itemize}

\subsection{Stage {\RomanNumeralCaps{1}}: Object Label Refiner} 
To efficiently and effectively evaluate a scene graph generation model, previous works~\cite{scenegraph:neural_motifs,scenegraph:IMP,scenegraph:graphrcnn,scenegraph:VTEN} have designed two experiments: SGCLS and PREDCLS (see details in Sec.~\ref{secEval}). All existing experimental results~\cite{scenegraph:neural_motifs,scenegraph:IMP,scenegraph:graphrcnn,scenegraph:VTEN} have shown that the scores (recall@20, 50, 100)  of PREDCLS are significantly higher than those of SGCLS by approximately 30\%. Both SGCLS and PREDCLS take ground truth boxes as inputs, but PREDCLS adopts the ground truth labels, while SGCLS takes the predicted object label. These results~\cite{scenegraph:neural_motifs} prove the importance of accurate object labels for scene graph generation. In other words, \textit{how to improve the prediction accuracy of object labels is the key to solving the scene graph generation problem.}

Therefore, in the second step, we aim to improve scene graph generation by refining the object labels generated by the Faster R-CNN network. The inputs of the Object Label Refiner are instance features, bounding boxes, and categories, and the outputs are refined object categories.
When generating object labels, the Faster R-CNN neither considers the global context~\cite{scenegraph:neural_motifs} nor object relations. Thus, we propose a relation regularized module for label refinement with a stack of Bi-LSTM~\cite{networks:mike_bilstm,networks:highway_lstm,networks:lstm} to capture the global context and a relation based graph convolution layer~\cite{networks:gcn_kipfs_semi,networks:gcn_rel,networks:gcn_srl} to make full use of object relationships.

\textbf{Relation Regularized Label Encoder.} 
\begin{figure}
	\begin{center}
		\includegraphics[width=0.9\linewidth]{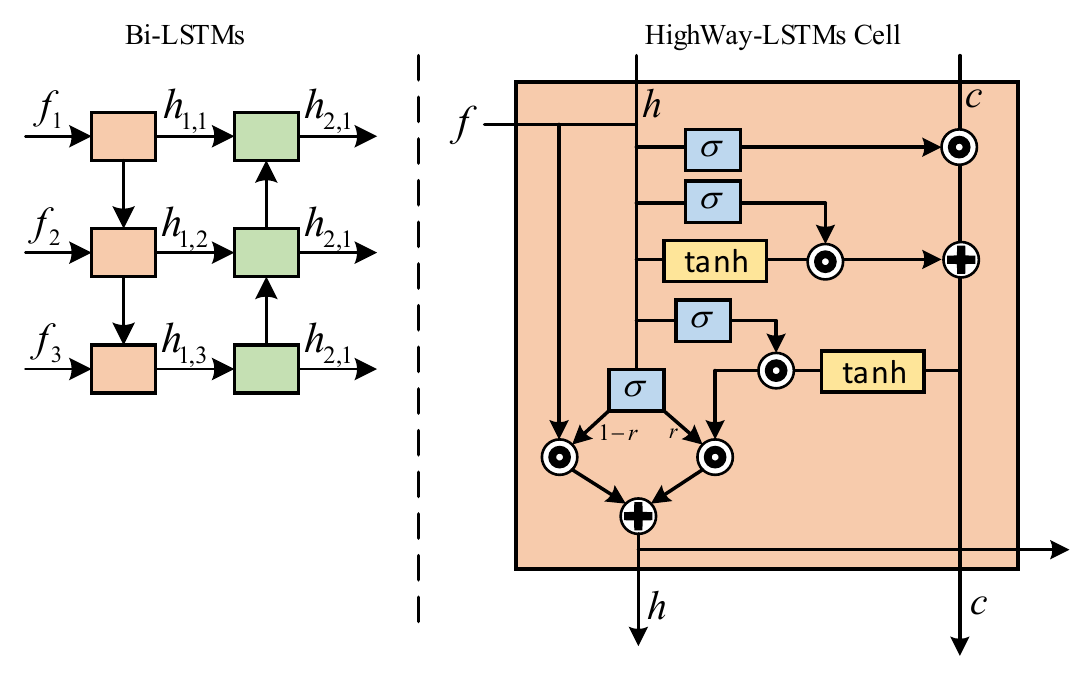}
	\end{center}
	\caption{The left figure shows the data stream of Bi-LSTMs. The right figure shows the cell of Bi-LSTMs with the highway connection. The operator $\circ$ represents the Hadamard product (also known as the element-wise product). The function $\sigma$ represents the sigmoid function. For clarity, we omit the subscript of the symbol in the right figure.}
	\label{fig:highwaylstm}
\end{figure}
In our work, we use deep bidirectional LSTMs to explore the global context. However, directly using deep LSTMs may exist the problems of training difficulty and slow convergence. In order to alleviate such issues, the highway LSTM~\cite{networks:highwaylstm_sr} was designed by connecting memory units of adjacent LSTM layers, and this structure~\cite{networks:highway_lstm,networks:highwaylstm_sr} has been used predominantly to achieve state-of-the-art results for semantic labeling, language modeling, etc. Therefore, we use stacked Bi-LSTMs with highway connections to encode object features $\textbf{F}$ to alleviate the problems caused by deep LSTMs:
\begin{equation}
\begin{array}{l}
i_{k,t} = \sigma(W^k_i [h_{k,t+\delta_k}, x_{k,t}] + b^k_i), \\
o_{k,t} = \sigma(W^k_o [h_{k,t+\delta_k}, x_{k,t}] + b^k_o), \\ 
f_{k,t} = \sigma(W^k_f [h_{k,t+\delta_k}, x_{k,t}] + b^k_f), \\
r_{k,t} = \sigma(W^k_r [h_{k,t+\delta_k}, x_{k,t}] + b^k_r), \\
g_{k,t} = \tanh(W^l_g [h_{k,t+\delta_k}, x_{k,t}] + b^k_g), \\

c_{k,t} = f_{k,t} \circ c_{k, t+\delta_k} + i_{k,t} \circ g_{k, t}, \\

h'_{k,t} = o_{k,t} \circ \tanh(c_{k,t}), \\
h_{k,t} = r_{k,t} \circ h'_{k,t} + (1 - r_{k,t} ) \circ W^k_h x_{k,t},
\end{array}
\label{equ.bi_hlstm} 
\end{equation}
where $x_{k,t}$ is the input to the $k$-th LSTM layer at time-step $t$, $W_*^*$ and $b_*^*$ are parameters, and $\delta_k$ indicates the direction of $k$-th  LSTM layer. The operator $\circ$ represents the Hadamard product (also known as the element-wise product). The function $\sigma$ represents the sigmoid function. Following He~\etal~\cite{networks:highway_lstm}, we set the inputs $x_{k,t}$ and $\delta_k$ of each LSTM layer as follows:
\begin{equation}
x_{k,t} = \left\{
\begin{array}{lr}
f_t & k=1 \\
h_{k-1,t} & k > 1 
\end{array},
\right.
\label{equ.inputs} 
\end{equation}
\begin{equation}
\begin{array}{l}
\delta_k = \left\{
\begin{array}{lr}
1 & k \bmod 2=0 \\
-1 & k \bmod 2=1
\end{array},
\right.
\end{array}
\label{equ.delta} 
\end{equation}
where $f_t$ is the $t$-th bounding box feature of $\textbf{F}$ extracted from the Faster R-CNN as mentioned in Sec.~\ref{OBSTP1}. The outputs of the highway LSTMs are denoted as  $\textbf{H} = \{h_1, h_2, ..., h_N\}$. The data stream of Bi-LSTMs with highway connection is shown as \ref{fig:highwaylstm}.

Graph Convolutional Networks (GCNs)~\cite{networks:gcn_kipfs_semi,networks:gcn_rel,networks:gcn_srl} produce an optimized node-level output, which encodes both original node features and the associations between data nodes. By using GCNs, our model can integrate the information of related objects to boost the performance of the label prediction.

With Faster R-CNN and stacked Bi-LSTMs, we can obtain a set of global features $\textbf{H} = \{h_1, h_2, ..., h_N\}$ corresponding to previously detected $N$ objects. By considering each object feature as a node, we take the global features $\textbf{H}$ to estimate an adjacency matrix $\textbf{A}^e = \{a^e_{1,1},...,a^e_{N,N}\}$, where $a^e_{i,j} \in \mathbb{R}^{1} $ represents whether a relationship exists from object $i$ to object $j$.

In the process of scene graph generation, an object label may act as a subject or an object in a scene graph. Thus we firstly adopt two fully connected layers to map $h_i$ into a subject space domain and an object space domain, respectively. 
\begin{equation}
\begin{array}{l}
h^s_i = FC^s_h(h_i),\\
h^o_i = FC^o_h(h_i).
\end{array}
\label{equ.adj_fc} 
\end{equation}
The output dimensions of the two fully connected layers are $D_f$. Next, we apply a simple and effective scoring function, DistMult~\cite{KB:distmult}, to compute affinity matrix scores:
\begin{equation}
\begin{array}{l}
a^e_{i,j} = \sigma((h^s_i \circ u_{i,j})^T W^a (h^o_j \circ u_{i,j}) + b^a_{i,j}),
\end{array}
\label{equ.adj_dismult} 
\end{equation}
where $W^a \in \mathbb{R}^{D_f \times D_f}$ is a diagonal parameter matrix that the model needs to learn. $b^a_{i,j} \in \mathbb{R}^{1}$ is a bias specific to the subject $i$ and object $j$ labels. Following~\cite{scenegraph:neural_motifs}, we initialize the bias with the frequency of the training set.  $\sigma$ is an activation function mapping the score ranging from $0$ to $1$. Besides, giving two objects it is difficult to determine the information flow direction from the object to the subject or vice versa. Therefore, the adjacency matrix is adjusted to form a symmetric matrix $\textbf{A}^s$ to solve this issue:
\begin{equation}
\begin{array}{l}
a^s_{i,j} = \left\{
             \begin{array}{lr}
				a^e_{i,j} & if \ a^e_{i,j} \geq a^e_{j,i} \\
				a^e_{j,i} & if \ a^e_{i,j} < a^e_{j,i} \\
				1 & if \ i=j
             \end{array}.
\right.
\end{array}
\label{equ.adj_ad} 
\end{equation}

With the generated symmetric matrix $\textbf{A}^s$, we integrate Bi-LSTMS with GCNs to obtain relational features $\textbf{O} = \{o_1,...,o_N\}$:
\begin{equation}
\begin{array}{l}
\textbf{O} = ReLU( \textbf{D}^s \textbf{A}^s \textbf{H} W^G),
\end{array}
\label{equ.gcns} 
\end{equation}
where $W^G$ is a parameter matrix.  $\textbf{D}^s = \{d^s_{1,1},..,d^s_{N,N}\}$ is a diagonal matrix:
\begin{equation}
\begin{array}{l}
 d^s_{i,j} =  \left\{
              \begin{array}{lr}
 				\frac{1}{{\sum\limits_{k = 1}^N {a_{i,k}^s} }} & if \ i=j \\
 				0 & if \ i \ne j
              \end{array}.
 \right.

\end{array}
\label{equ.dnor} 
\end{equation}
Next, we concatenate the global context $h_i$ and the relational feature $o_i$ to form the final output ${{o'}_i}$:
\begin{equation}
\begin{array}{l}
{{o'}_i}={[o_i,h_i]}.

\end{array}
\label{equ.dnor2} 
\end{equation}

For simplicity, we define the whole Relation Regularized encoding process as:
\begin{equation}
\begin{array}{l}
 \left\{ {\textbf{A}^e}, \textbf{O}' \right\}= {\rm{R2\_Encoder}}\left( \textbf{F} \right|W^{o}),
\end{array}
\label{equ.dnor3} 
\end{equation}
where $W^{o}$ is the parameter in the Relation Regularized module.

\textbf{Label Decoder.} Finally, we use an LSTM layer with Highway gate to decode $\textbf{O}'$. After refining the previous generated initial object labels $\textbf{L}$, we obtain the refined object label ${\textbf{L}^d}$: 

\begin{equation}
\begin{array}{l}
q^d_i = LSTM([l^d_{i-1}, q^d_{i-1}, {{o'}_i}]), \\ 
l^d_i = argmax(W q^d_i + l_i),
\end{array}
\label{equ.dec_lstm} 
\end{equation}
where $l_i$ is the prior label distribution from Faster R-CNN as mentioned in Sec.~\ref{OBSTP1}. $q^d_i$ is the hidden state of LSTM. $l^d_i$ is $i$-th the refined object labels ${\textbf{L}^d}$. In addition, we set $\left\langle {BOS} \right\rangle$ as the start signal for the decoding process. 
 
\subsection{Stage {\RomanNumeralCaps{2}}: Object Relationship Generator}
The Object Relationship Generator predicts the relationship predicate $\textbf{R}$ between instances with instance features $\textbf{O}'$ from Eq.~\ref{equ.dnor3} and refined labels $\textbf{L}$ from Eq.~\ref{equ.dec_lstm}. 

\textbf{Relation Regularized Relationship Encoder.} With Faster R-CNN and our proposed relation regularized label generation module, we can transfer an image into a set of optimized instance features $\textbf{O}'$ and refined labels ${\textbf{L}^d}$. Next, we apply our proposed relation regularized encoder to facilitate the relationship prediction process:
\begin{equation}
\begin{array}{l}
\left\{ {\textbf{A}^r}, \textbf{Z} \right\}= {\rm{R2\_Encoder}}\left( {[\textbf{O}',{W^L}{\textbf{L}^d}]} \right|W^{z}),

\end{array}
\label{equ.encodr2} 
\end{equation}
where $W^l$ is the embedding parameter and is initialized by Glove~\cite{word2vec:glove}, and $\textbf{Z}=\left\{ {{z_1}, \cdots ,{z_N}} \right\}$ is the output feature of R2-Encoder in the Relationship Generator phase. $\textbf{Z}$ is similar to  $\textbf{O}'$  in Eq.~\ref{equ.dnor3} and contains the global and relational features in the Relationship Generator.

\textbf{Relationship Decoder.} Next, we obtain the object relationship by firstly mapping ${z_i}$ into two feature space domains: subject domain and object domain, $z^s_i$ and $z^o_i$ respectively:
\begin{equation}
\begin{array}{l}
z^s_i = FC^s_h(z_i),\\
z^o_i = FC^o_h(z_i). 
\end{array}
\label{equ.adjfc2} 
\end{equation}
We then employ the DisMult score function to compute the object relations:
\begin{equation}
\begin{array}{l}
r'_{m, i, j} = (z^s_i \circ u_{i,j})^T W^r_m (z^o_j \circ u_{i,j}) + b^r_{m, i,j},
\end{array}
\label{equ.rel_dismult} 
\end{equation}
where $r'_{m, i, j}$ is the probability that the object $i$ and object $j$ belong to the relationship $m$. $W^r_m$ is a diagonal parameter matrix that needs to be learned.  $b^r_{m,i,j}$ is the frequency bias mentioned in~\cite{scenegraph:neural_motifs}. Finally, we use a softmax function to obtain the final relation score ranging from $0$ to $1$:
\begin{equation}
\begin{array}{l}
r_{m, i, j} = 

\frac{{{e^{r'_{m,i,j}}}}}{{\sum\limits_{m = 1}^{{D_r}} {{e^{r'_{m,i,j}}}} }}
\end{array},
\label{equ.rel_softmax} 
\end{equation}
where $D_r$ is the number of relationship categories in the dataset. Now we can get the final relationships $\textbf{R} = \{r_{1,1,1},...,r_{D_r,N,N}\}$.

\subsection{Loss Functions}
\label{sec.lossfunc}
In this section, we first describe two objective functions. The first one is the label prediction loss $\mathcal{L}_1$ for refining the object labels, while the second loss is 
relation regularized loss (R2-loss $\mathcal{L}_2$) for learning the first adjacency matrix:
\begin{equation}
\begin{array}{l}
{\mathcal{L}_1} = \rm{Cross\_Entropy}\left( {{\textbf{L}^d},{\textbf{L}^g}} \right), \\
{\mathcal{L}_2} = \rm{Cross\_Entropy}\left( {{\textbf{A}^e},{\textbf{A}^g}} \right),
\end{array}
\label{equ.loss_func1} 
\end{equation}
where ${\textbf{L}^d}$ is the output from Eq.~\ref{equ.dec_lstm}, and $\textbf{L}^g$ is the ground truth object labels. ${\textbf{A}^e}$ is obtained by Eq.~\ref{equ.adj_dismult}. ${\textbf{A}^g}$ is the ground truth adjacency matrix, which is used to indicate whether there is a relationship between two entities, \ie 0 or 1.

To learn the parameters for generating relations between objects, we describe two object functions to control our model:
\begin{equation}
\begin{array}{l}
{\mathcal{L}_3} = \rm{Cross\_Entropy}\left( {{\textbf{A}^r},{\textbf{A}^g}} \right),\\
{\mathcal{L}_4} = \rm{Cross\_Entropy}\left( {{\textbf{R}},{\textbf{R}^g}} \right), \\

\end{array}
\label{equ.loss_func2} 
\end{equation}
where ${\mathcal{L}_3}$ is another relation regularized loss at the relationship generation phase, and ${\mathcal{L}_4}$ is the object relationship loss (R2-loss). The proposed final objective function of our proposed a relation regularized model is defined as the sum of ${\mathcal{L}_1}$, ${\mathcal{L}_2}$, ${\mathcal{L}_3}$ and ${\mathcal{L}_4}$. 

\section{Experiments}
In this section, we evaluate our model on the cleaned Visual Genome dataset~\cite{scenegraph:visual_genome}. Some experiments are conducted to test the role of the major components and to compare our model with the previous methods.
\subsection{Experimental Setting: Dataset}
Krishna~\etal~\cite{scenegraph:visual_genome} collected and officially released a knowledge base to connect structured images concept to languages, namely Visual Genome dataset. They have collected more than $100$k images. On average, each image contains $38$ objects and $22$ pairs of object relations. This is an ideal candidate dataset for evaluating scene graph generation models. However, some annotations are ambiguous and may lead to predicting errors. Many cleaning approaches have been proposed, such as~\cite{scenegraph:denscap,scenegraph:IMP,scenegraph:drnet}.  In particular, Xu~\etal~\cite{scenegraph:IMP} proposed a cleaning strategy to remove ambiguous annotations. This strategy has been widely adopted by previous scene graph generation methods, such as~\cite{scenegraph:IMP,scenegraph:VRD_LP,scenegraph:PGAE,scenegraph:neural_motifs,scenegraph:graphrcnn}. After cleaning, each image, on average, contains approximately $12$ objects and $6$ pairs of relationships. In total, the cleaned Visual Genome dataset contains  $150$ object categories and $50$ object relation classes. Moreover, we follow Xu~\etal~\cite{scenegraph:IMP} to divide the cleaned dataset into two subsets: $70$\% training and $30$\% testing. Next, we follow~\cite{scenegraph:IMP,scenegraph:VRD_LP,scenegraph:PGAE,scenegraph:neural_motifs,scenegraph:graphrcnn} to select $5$k images from the training dataset as the validation set.

\subsection{Experimental Setting: Implementation Details}
\label{sec.imp_det}
For object detection step, we follow the previous works~\cite{scenegraph:IMP,scenegraph:neural_motifs} to adopt the Faster R-CNN model as the object detector to generate a set of bounding boxes and their corresponding features and labels. More specifically, previous works~\cite{scenegraph:IMP,scenegraph:PGAE,scenegraph:neural_motifs, scenegraph:graphrcnn,scenegraph:GPI} adopt the VGG16 network as the backbone of pre-trained Faster R-CNN. For a fair comparison, we also adopt the VGG16 based Faster R-CNN network, which is pre-trained on the Visual Genome object dataset by Zellers~\etal~\cite{scenegraph:neural_motifs}. For the VGG16 based Faster R-CNN network, we obtain the region features from the output feature map of the second fully connected layer. 

By using ResNet101 to extract deeper features, previous work~\cite{obj_det:faster_rcnn} has proven that ResNet101-based Faster R-CNN can perform better than VGG16-based Faster R-CNN on the object detection task. Better region features and more accurate object labels may be conducive to object relation generation. Therefore, we first take the ResNet101~\cite{classify:resnet} as the backbone of the Faster R-CNN and then train the ResNet101 based Faster R-CNN on the training set of the cleaned Visual Genome dataset. More specifically, the detector is trained on a Titan Xp ($12$ GB) with the SGD optimizer. The learning rate is set as $1 \times 10^{-3}$ and the batch size is set as $6$. We compare the VGG16 based Faster R-CNN model with the ResNet101 based one on the cleaned test dataset in ablation studies. In addition, we take the $pool\_5$ layer of ResNet101 as the output feature map for extracting region features.

For the second step, label refinement, our relation regularized encoder requires the input regions to be sorted in an order. Therefore, we sort the object features from left to right as our encoder's inputs. A previous study~\cite{scenegraph:neural_motifs} has proven that the order of the Bi-LSTMs inputs has little effect to extract global context based region features. For the first relation regularized encoder, we set the number of Bi-LSTM layers as $2$. For the second relation regularized encoder of relation generation, we set the number of Bi-LSTM layers as $4$. 

We first train the detector (Faster R-CNN) on the Visual Genome dataset as mentioned above, and then freeze the parameters of Faster R-CNN when processing the scene graph generation.
When training our model on the scene graph classification (SGCLS) and the predicate classification (PREDCLS), we follow previous works~\cite{scenegraph:PGAE,scenegraph:IMP} to use the ground truth bounding boxes for refining object labels and generating object relations. In order to learn two object relation matrices $A^{e}$ and $A^{r}$, we sample twice as many positive samples as negative samples. For relation $\textbf{R}$, we utilize the same number of positive and negative samples. Moreover, we choose SGD with momentum~\cite{opt:sgd} as our optimizer. The learning rate and batch size are set to $2 \times 10^{-2}$ and $24$, respectively. When training our model on the scene graph detection (SGDET), we fine-tune our model following the strategy mentioned in~\cite{scenegraph:neural_motifs} for fairness. We sample 256 Region of Interests (RoIs) in each image. After the label decoder in the object label refiner, we use the per-category Non-Maximum Suppression (NMS) to filter redundant objects.

Our codes are implemented in python. We use Pytorch to build our model. All experiments are conducted on a Ubuntu server with $2$ Titan Xps (12 GB), $4$ Intel(R) Xeon(R) E5-2650 CPUs and $256$ GB RAM.
\begin{table}[]
	\centering	
	\caption{The differences among three tasks: PREDCLS, SGCLS and SGDET. Check marks indicate what information a task needs to predict. }
	\scalebox{0.9}{
\begin{tabular}{c|c|c|c}
\hline
\multicolumn{1}{c|}{\multirow{2}{*}{Tasks}} & \multicolumn{3}{c}{Required Prediction}                    \\ \cline{2-4} 
\multicolumn{1}{c|}{}                       & Relationship Predicate & Object Label & Object Bounding Box \\ \hline
PREDCLS                                      & $\checkmark$           &              &                     \\ \hline
SGCLS                                        & $\checkmark$           & $\checkmark$ &                     \\ \hline
SGDET                                        & $\checkmark$           & $\checkmark$ & $\checkmark$        \\ \hline
\end{tabular}}
\label{tab.diff_tasks}
\end{table}
\begin{table}[]
	\centering	
	\caption{The role of relation regularized label refiner. The results are obtained on the validation dataset.}
	\begin{tabular}{c|c|c|c|l}
		\hline
		\multirow{2}{*}{R2 Variants} & \multicolumn{4}{c}{SGCLS}    \\ \cline{2-5} 
		& R@20 & R@50 & R@100 & obj acc \\ \hline
		w/o refiner        & 33.3 & 39.1  & 41.1    & 71.1    \\ \hline
        w/o prior labels  & 38.7 & 42.1  & 42.9    & 71.5    \\ \hline  \hline
		All       & 39.9 & 42.6 & 43.3  & 72.3    \\ \hline

	\end{tabular}

	\label{Tab.priorlabel}
\end{table}

\subsection{Experimental Setting: Evaluation Strategy}
\label{secEval}

Following previous works~\cite{scenegraph:IMP,scenegraph:graphrcnn,scenegraph:PGAE}, we evaluate our model with three experimental setups: 1) The \textbf{PREDCLS} (predicate classification) allows models to take the ground truth bounding boxes and the ground truth object labels as inputs. 2) The \textbf{SGCLS} (scene graph classification) allows models to take the ground truth bounding boxes as inputs. And 3) The \textbf{SGDET} (scene graph detection), which requires a model to take an image as inputs and then predict object bounding boxes, object labels, and object relations. Tab.~\ref{tab.diff_tasks} shows what information a task needs to predict. Since the PREDCLS task allows models to take the ground truth bounding boxes and the ground truth object labels as inputs, it only requires models to predict relational predicates. However, the SGCLS task requires models to predict relationship predicates and object labels, and the SGDET task requires models to predict relationship predicates, object labels and object bounding boxes. In other words, SGDET is the most difficult one, while PREDCLS is the easiest one. Moreover, in order to provide a more consistent comparison, we also report the corresponding results of the three setups without scene graph constraints. 

The object relations in the Visual Genome dataset are sparse, thus using the mean average precision (mAP) as the evaluation metric would falsely punish positive predictions on unannotated relations. As a result, we follow previous works~\cite{scenegraph:VRD_LP,scenegraph:PGAE,scenegraph:IMP,scenegraph:neural_motifs} to evaluate our model with recall@K (i.e., R@K). More specifically, R@K describes the proportion of ground truth triplets (i.e., obj1-relation-obj2) in the top $K$ predicted triplets. The $K$ is set as $20$, $50$ and $100$. Besides, for the \textbf{SGDET} task, if the object has at least $0.5$ IoU overlapping with the ground truth box, it is considered as correctly detected. 

\subsection{Results and Analysis: Speed of R2-Net}

Regarding the speed of our model, in the training phase, each epoch (about $57$k images) takes $40$ minutes for the SGCLS task and $210$ minutes for SGDET. In the testing phase, our model takes about $0.12$ seconds to parse a single image for SGCLS and PREDCLS, and $0.33$ seconds for SGDET. These results are obtained under the experimental environment mentioned in Sec.~\ref{sec.imp_det}.
 
\subsection{Results and Analysis: Ablation Study}
In order to deeply analyze the proposed approach and demonstrate its effectiveness, we present an extensive ablation study on the Visual Genome dataset by considering different variants of the proposed R2-Net to evaluate its major components. In this section, the experimental results are obtained from the validation dataset and we choose ResNet101 based Faster R-CNN network as our object detector.

\begin{table}[]
	\centering
	\caption{The role of Bi-LSTMs and GCNS in R2-Net.The results are obtained on the validation dataset.}
	\begin{tabular}{c|c|c|c|c}
		\hline
		\multirow{2}{*}{R2 Variants} & \multicolumn{4}{c}{SGCLS}       \\ \cline{2-5} 
	& R@20 & R@50 & R@100 & obj acc \\ \hline
		w/o Bi-LSTM1           & 39.1 & 41.9  & 42.7    & 71.8    \\ \hline
		w/o Bi-LSTM2          & 39.7 & 42.3  & 43.1    & 72.2    \\ \hline \hline
	    w/o GCN1                 & 39.0 & 41.8  & 42.5    & 71.8    \\ \hline
	    w/o GCN2                & 39.4 & 42.4  & 43.2    & 72.1    \\ \hline  \hline
	    	
	   w/o GCNs    & 38.3 & 42.0  & 43.0    & 72.2    \\ \hline
	   w/o Bi-LSTMs  & 39.2 & 42.0  & 42.8    & 71.9    \\ \hline  \hline
	   w/o R2-Loss  & 38.4 & 42.1  & 43.0    & 71.9    \\ \hline  \hline
	    All       & 39.9 & 42.6 & 43.3  & 72.3    \\ \hline      
	\end{tabular}

	\label{Tab.BILSTMS}
\end{table}
\begin{table}[]
	
	\begin{center}
			\caption{The role of two deep learning features: Vgg16 and ResNet101. The results are obtained on the testing dataset.}
\resizebox{0.5\textwidth}{!}{
			\begin{tabular}{c|c|c|c|c|c|c}
				\hline
				\multirow{2}{*}{Model} & \multicolumn{3}{c|}{SGCLS}                    & \multicolumn{3}{c}{PREDCLS}                  \\ \cline{2-7} 
				& R@20          & R@50          & R@100         & R@20          & R@50          & R@100         \\ \hline
				Motifnet(VGG)~\cite{scenegraph:neural_motifs}          & 32.9          & 35.8          & 36.5          & 58.5          & 65.2          & 67.1          \\ \hline
				Motifnet(ResNet)          & 33.1          & 36.0          & 36.7          & 58.5          & 65.0          & 66.8          \\ \hline
				R2-Net(VGG)     & 33.5          & 36.5          & 37.3          & \textbf{59.2} & \textbf{65.9} & \textbf{67.8} \\ \hline
				R2-Net(ResNet)      & \textbf{34.5} & \textbf{37.5} & \textbf{38.3} & 59.0          & 65.5          & 67.3          \\ \hline
		\end{tabular}}

		\label{tab.feat}
	\end{center}
\end{table}
\begin{table}[]
\begin{center}
\caption{Performance per predicate predicate (top ten on left, bottom ten on right).}
\begin{tabular}{c|c|c|c}
\hline
Predicate  & Recall & Predicate  & Recall \\ \hline
wearing    & 96.2   & across     & 13.7   \\
on         & 93.4   & painted on & 11.6   \\
riding     & 92.8   & against    & 9.7    \\
has        & 91.9   & between    & 9.5    \\
of         & 91.5   & mounted on & 8.9    \\
wears      & 86.9   & growing on & 3.4    \\
holding    & 86.1   & playing    & 1.0    \\
in         & 82.6   & from       & 0.0    \\
walking on & 81.9   & says       & 0.0    \\
sitting on & 78.9   & flying in  & 0.0   \\ \hline
\end{tabular}
\label{tab.pre}
\end{center}
\end{table}

\begin{table*}[]
	\begin{center}		
	\caption{Comparison with other methods. The results are obtained on the test dataset. The results of IMP+ are reproduced in~\cite{scenegraph:neural_motifs}. The supervised learning strategy is denoted as SL. The Reinforcement Learning strategy is denoted as RL.}	
		\begin{tabular}{c|c|c|c|c|c|c|c|c|c|c}
			\hline
			\multicolumn{1}{l|}{\multirow{2}{*}{Learning Strategy}} &
			\multirow{2}{*}{Model}  & \multicolumn{3}{c|}{SGDET}                    & \multicolumn{3}{c|}{SGCLS}                    & \multicolumn{3}{c}{PREDCLS}                  \\ \cline{3-11}
			\multicolumn{1}{l|}{} 
			 & & R@20          & R@50          & R@100         & R@20          & R@50          & R@100         & R@20          & R@50          & R@100         \\ \hline
			\multirow{10}{*}{SL} 
			& VRD~\cite{scenegraph:VRD_LP}          &   -        & 0.3          & 0.5          &    -       & 11.8          & 14.1          &   -        & 27.9         & 35.0          \\ \cline{2-11}
			& IMP~\cite{scenegraph:IMP}        & -         & 3.4          & 4.2          & -          & 21.7          & 24.4          & -          & 44.8          & 53.0          \\ \cline{2-11}
			& IMP+~\cite{scenegraph:IMP}        & 14.6          & 20.7          & 24.5          & 31.7          & 34.6          & 35.4          & 52.7          & 59.3          & 61.3          \\ \cline{2-11}
			& TFR~\cite{scenegraph:TFR}        & 3.4         & 4.8          & 6.0          & 19.6          & 24.3          & 26.6          & 40.1          & 51.9          & 58.3          \\ \cline{2-11}
			& AE~\cite{scenegraph:PGAE}          & 6.5           & 8.1           & 8.2           & 18.2          & 21.8          & 22.6          & 47.9          & 54.1          & 55.4          \\ \cline{2-11}
			& FREQ+OVERLAP~\cite{scenegraph:neural_motifs}             & 20.1          & 26.2          & 30.1          & 29.3          & 32.3          & 32.9          & 53.6          & 60.6          & 62.2          \\ \cline{2-11}
			& Graph R-CNN~\cite{scenegraph:graphrcnn}             & -             & 11.4          & 13.7          & -             & 29.6          & 31.6          & -             & 54.2          & 59.1          \\ \cline{2-11}

			& Motifnet~\cite{scenegraph:neural_motifs}                & 21.4          & 27.2          & 30.3          & 32.9          & 35.8          & 36.5          & 58.5          & 65.2          & 67.1          \\ \cline{2-11} 
			\cline{2-11}
			& R2-Net (w/o GCN1) & \textbf{23.1} & \textbf{29.4} & \textbf{33.0} & 33.6          & 36.5          & 37.3          & 58.8          & \textbf{65.6}          & \textbf{67.4}          \\ \cline{2-11}
			& R2-Net           & 21.6          & 27.5          & 31.3          & \textbf{34.5} & \textbf{37.5} & \textbf{38.3} & \textbf{59.0} & 65.5 & 67.3 
			\\ \hline \hline
			\multirow{2}{*}{SL+RL} 
			& VCTREE~\cite{scenegraph:treelstm}                & 22.0          & 27.9          & 31.3          & 35.2          & 38.1          & 38.8          & 60.1          & 66.4          & 68.1          \\ \cline{2-11}
			& CMAT~\cite{scenegraph:cmat}                & 22.1          & 27.9          & 31.2          & 35.9          & 39.0          & 39.8          & 60.2          & 66.4          & 68.1          \\ \hline
		\end{tabular}
		\label{Comp1}
	\end{center}
\end{table*}

\begin{table*}[]
	\begin{center}
			\caption{Comparison with other methods. Performance is computed without graph constraints. The results are obtained on the test dataset. The supervised learning strategy is denoted as SL. The Reinforcement Learning strategy is denoted as RL.}
		\begin{tabular}{c|c|c|c|c|c|c|c}
			\hline
			\multicolumn{1}{l|}{\multirow{2}{*}{Learning Strategy}} &
			\multirow{2}{*}{Model} &
 			\multicolumn{2}{c|}{SGDET}    & \multicolumn{2}{c|}{SGCLS}    & \multicolumn{2}{c}{PREDCLS}  \\ \cline{3-8} 
 			\multicolumn{1}{l|}{} 
		  &	& R@50          & R@100         & R@50          & R@100         & R@50          & R@100         \\ \hline
			\multirow{6}{*}{SL}  & IMP~\cite{scenegraph:IMP}                    & 22.0          & 27.4          & 43.4          & 47.2          & 75.2          & 83.6          \\ \cline{2-8}
			& AE~\cite{scenegraph:PGAE}                       & 9.7           & 11.3          & 26.5          & 30.0          & 68.0          & 75.2          \\ \cline{2-8}
			
			&  FREQ+OVERLAP~\cite{scenegraph:neural_motifs}                        & 28.6          & 34.4          & 39.0          & 43.4          & 75.7          & 82.9          \\ \cline{2-8}
			&  Motifnet~\cite{scenegraph:neural_motifs}                           & 30.5          & 35.8          & 44.5          & 47.7          & 81.1          & 88.3          \\ \cline{2-8} 
			&  R2-Net (w/o GCN1)               & \textbf{32.9} & \textbf{38.6} & 45.4          & 48.8          & 81.0          & \textbf{88.4} \\ \cline{2-8}
			& R2-Net                    & 30.4          & 36.0          & \textbf{46.6} & \textbf{49.9} & \textbf{81.3} & 88.2          \\ 
			\hline
			\hline
			\multirow{1}{*}{SL+RL} 
			& CMAT~\cite{scenegraph:cmat}                
			& 31.6          & 36.8          & 48.6          
			& 52.0          & 83.2          & 90.1            \\ \hline 
		\end{tabular}

		\label{Comp2}
	\end{center}
\end{table*}
\textbf{Effect of Object Label Refiner}
We consider three R2-Net variants: 1) \textit{w/o refiner} where R2-Net performs the scene graph generation with only two components. The object detector is followed by an object relation generator; 2) \textit{w/o prior labels} where the object label refiner without the prior labels mentioned in Eq.~\ref{equ.dec_lstm}; 3) \textit{All} where R2-Net performs the task with all three components: object detector, object label refiner and object relation generator. We carry out the experiments, as shown in Tab.~\ref{Tab.priorlabel}. It can be observed that \textit{all} significantly performs better than \textit{w/o refiner} on the \textbf{SGCLS} task with an increase of $6.6$\% on R@20, $3.5$\% on R@50 and $2.2$\% on R@100, while slightly outperforms the label detection by $1.2$\% in terms of accuracy. The experimental results prove that our proposed label refiner is able to extract more representative features for both object label classification and scene graph generation. Compared with \textit{w/o prior labels}, using prior labels can improve $1.2$\% on SGCLS (R@20) and $0.8$\% on object accuracy. Therefore, we use prior labels in the object label refiner to make the model easier to learn.

\textbf{Effect of Relation Regularized Encoder}
Our R2-Net consists of two relation regularized encoders. The first one is proposed for supporting the object label refiner, while the second one is designed for facilitating the object relationship generation. Both encoders consist of two components: 1) highway based Bi-LSTMS for capturing the global context and 2) object relation based graph convolutional layers for learning object relations. In order to deeply evaluate the effect of Bi-LSTMs and relation regularized GCNs of the two encoders, we design a set of R2-Net variants by removing the Bi-LSTMs of the first encoder (\textit{w/o Bi-LSTM1}), removing the Bi-LSTMs of the second encoder (\textit{w/o Bi-LSTM2}), removing the GCNs of the first encoder (\textit{w/o GCN1}), removing the GCNs of the second encoder (\textit{w/o GCN2}), removing all GCNs of the two encoders (\textit{w/o GCNs}), removing all Bi-LSTMs of the two encoders (\textit{w/o Bi-LSTMs}), and the full R2-Net (\textit{All}). 

The experimental results are shown in Tab.~\ref{Tab.BILSTMS} and we have the following observations: 
1) compared with \textit{All}, \textit{w/o Bi-LSTM1} decreases the R2-Net performance more by 0.8\% R@20, 0.7\% R@50 and 0.6\% R@100. This demonstrates the effectiveness of the proposed Bi-LSTM1 for encoding global context; 
2) removing either GCNs (GCN1 or GCN2) could lead to a drop in both tasks, scene graph generation and label prediction. However, without GCN1, the performance scores drop dramatically; 
3) compared with other R2-Net variants, \textit{w/o GCNs} obtains the lowest scores for R@20 with 38.3\%, which is 1.6\% lower than R2-Net. This demonstrates the superiority of the proposed GCNs for relation graph generation; 
4) compared with \textit{w/o R2-loss} (mentioned in Eq.~\ref{equ.loss_func1} and Eq.~\ref{equ.loss_func2}), using the relation regularized loss can bring additional supervision information to the model and improve the robustness of the model; 
5) compared with all variants, R2-Net (\textit{All}) performs best on all tasks; and 6) another fact is that \textit{w/o GCN2} and \textit{w/o Bi-LSTM2} have little effects on the SGCLS task. This shows that the prediction of labels is very important on the SGCLS task because \textit{GCN2} and \textit{Bi-LSTM2} are at the relationship generation phase. This further proves that each component is helpful and contributes to the final object detection and scene graph generation.

\begin{figure*}
	\begin{center}
		\includegraphics[width=0.85\linewidth]{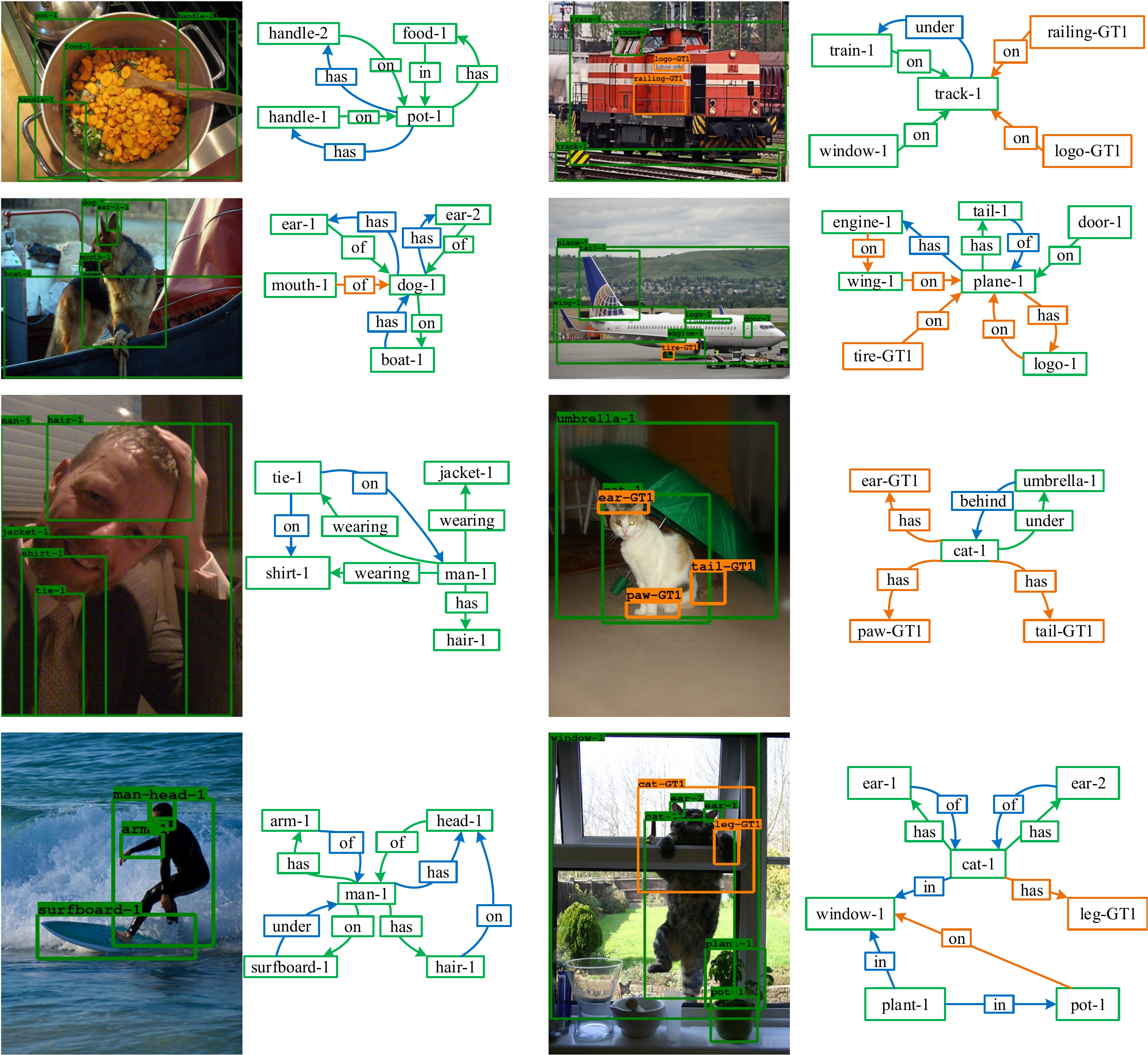}
	\end{center}
	\caption{Some qualitative examples produced by our R2-Net (w/o GCN1) on the \textbf{SGDET}. Predicted true positive boxes are marked with green (IOU$>0.5$). Orange boxes are ground truth boxes but not detected. For simplicity, we demonstrate the top 20 object relations or edges at the R@20 setting. Predicted true positives relations are marked with green arrows, false negative relations are marked with orange arrows, and false positive relations are marked with blue arrows.}
	\label{Fig:qualitative}
\end{figure*}
\begin{figure*}
	\begin{center}
		\includegraphics[width=0.85\linewidth]{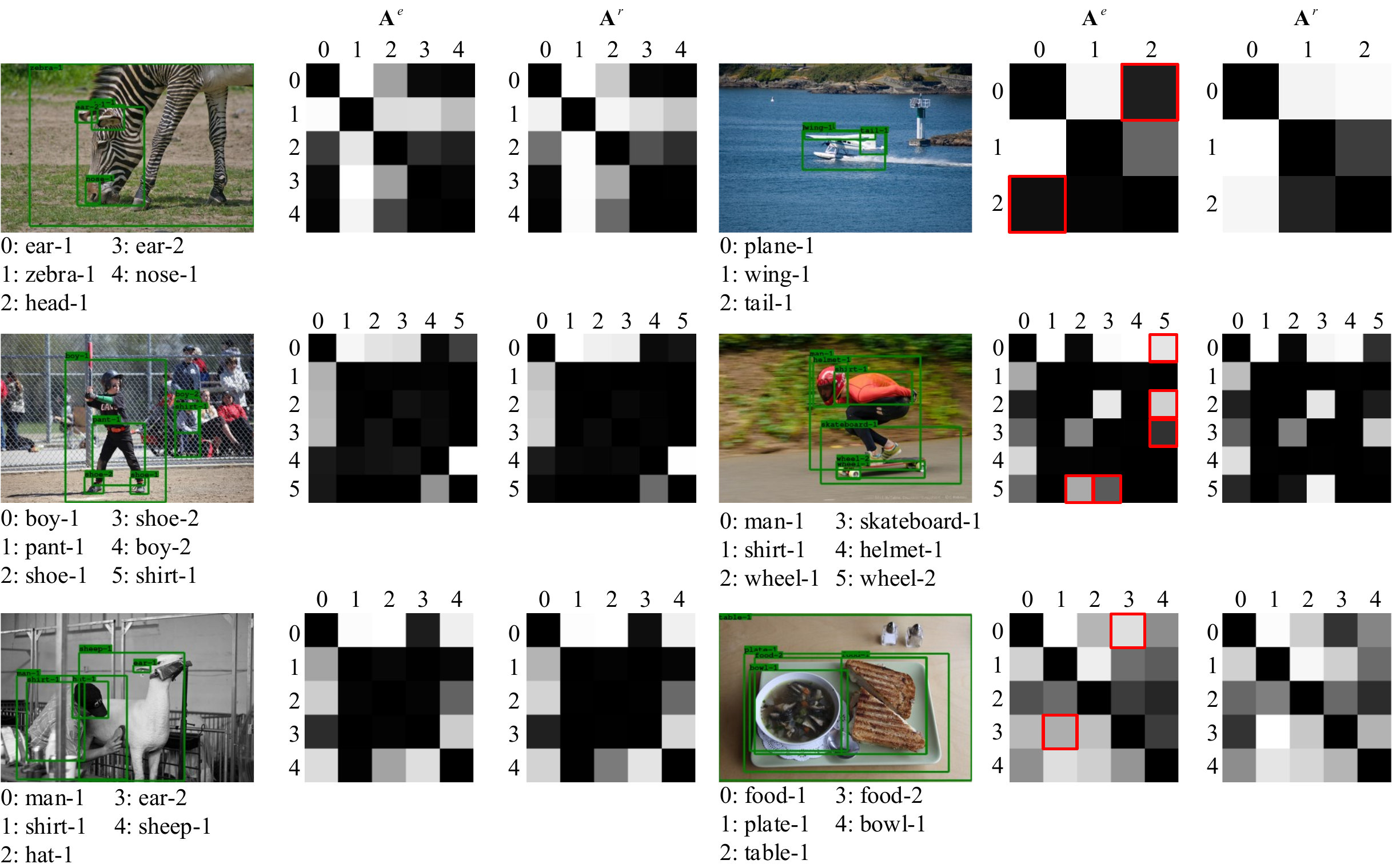}
	\end{center}
	\caption{Some affinity matrix examples produced by R2-Net on \textbf{SGCLS}. The small gray square indicates whether there is a relationship between two objects. The lighter the color, the greater the probability. The red bounding boxes indicate visible differences between $\textbf{A}^e$ and $\textbf{A}^r$ mentioned in Eq.~\ref{equ.dnor3} and Eq.~\ref{equ.encodr2}.}
	\label{Fig:qualitative_adj}
\end{figure*}
\textbf{Effect of CNN Backbones}
\label{DEEPFE}
As mentioned in Sec.~\ref{sec.imp_det}, the different backbones of the object detector can make the model produce different performances, so we compare the different backbones (VGG16 and ResNet101) in this section.

We first compare the performances of these two backbones on the object detection task. The ResNet101-based Faster R-CNN performs better than the VGG16-based model ($22.8$ mAP vs $20.0$ mAP at $0.5$ IoU) on the Visual Genome dataset. This result fully demonstrates the advantages of ResNet.

We further evaluate the effect of CNN backbones of the detector on the task of \textbf{SGCLS} and \textbf{PREDCLS}.
Here we choose to run the best previous method Motifnet and our proposed R2-Net to obtain the results on both tasks. The experimental results are shown in Tab.~\ref{tab.feat}. More specifically, the experimental results demonstrate that on the \textbf{SGCLS} task, the ResNet101-based model is obviously better than the corresponding VGG16-based model. Interestingly on the \textbf{PREDCLS} task, the VGG16-based models achieve higher scores than the corresponding ResNet101-based ones. In addition, with the same evaluation metrics and the same backbones, the performance score obtained on the \textbf{PREDCLS} task is around twice higher than the score reached from the \textbf{SGCLS}. From the experimental results, we can conclude that scene graph generation is largely depending on the accuracy of the object label prediction, while the label prediction accuracy is mostly relying on the region features. The more representative the region feature is the more accurate the object label prediction is. Moreover, from the experimental results, we can see that higher level features have only a slight effect on object label relation generation. More importantly, our R2-models significantly outperform the previous best method Motifnet on both two tasks, reaching the new state-of-the-art.

\textbf{Which of the predicates perform better/worse?} We investigate the performance of each predicate on the PREDCLS task (R@100) without graph constraints. In Tab.~\ref{tab.pre}, the top ten predicates with the highest score are shown on the left side and the bottom ten predicates with the lowest score are shown on the right side. By analyzing the dataset, we find that the samples of the top ten categories account for $78.9$\% of the total dataset, while the samples of the bottom ten categories account for only $1.3$\% of the total dataset. Therefore, the imbalance of the dataset is the main reason for the huge difference in predicate scores.

\subsection{Results and Analysis: Comparison with State-of-the-art Methods}
In this section, we compare our proposed R2-Net with several state-of-the-art methods, including Visual Relation Detection (VRD)~\cite{scenegraph:VRD_LP}, Iterative Message Passing (IMP)~\cite{scenegraph:IMP}, Tensorize Factorize Regularize (TFR)~\cite{scenegraph:TFR}, Associative Embedding (AE)~\cite{scenegraph:PGAE}, Graph R-CNN~\cite{scenegraph:graphrcnn}, FREQ+OVERLAP ~\cite{scenegraph:neural_motifs}, and Motifnet~\cite{scenegraph:neural_motifs}.  We cannot compare our method with Factorisable Net~\cite{scenegraph:factnet}, DR-Net~\cite{scenegraph:drnet}, and MSDN~\cite{scenegraph:denscap}, because the approaches for data cleaning and splitting are different. Besides, Factorisable Net~\cite{scenegraph:factnet} has proven that the more bounding boxes generated, the better performances of scene graph generation are. However, the more bounding boxes are chosen, the more complex the computation is. 
Motivated by this, for the task of  \textbf{SGDET} we choose the top $64$ regions detected by the Faster R-CNN, following previous works~\cite{scenegraph:neural_motifs,scenegraph:IMP,scenegraph:TFR}. To fully evaluate our method, we apply two evaluation strategies: with and without graph constraints. The experimental results are shown in Tab.~\ref{Comp1} (with graph constraints) and Tab.~\ref{Comp2} (without graph constraints). From them, we can see that with two experimental settings, our R2-Net performs the best on all three tasks with the supervised learning strategy, which confirms the effectiveness of our proposed method. Compared with the methods using reinforcement learning, our R2-Net (w/o GCN1) still achieves better results on the SGDET task. Because it is not practical to provide the ground truths of object bounding boxes or categories for models in real life, the SGDET task has more practical value than SGCLS or PREDCLS. Therefore, our method is superior to VCTREE and CMAT in some aspects.

More specifically, as shown in Tab.~\ref{Comp1}, R2-Net without graph convolutional layers (GCN1) of the first relation regularized encoder achieves the highest scores for the \textbf{SGDET} task, reaching 23.1\% on R@20, 29.4\% on R@50, and 33.0\% on R@100, which is significantly higher than the R2-Net. However, for the task of \textbf{SGCLS}, the performance result is opposite. Also, with the same metric and model, the score of the task \textbf{SGCLS} is around 10\% higher than the score of the \textbf{SGDET} task. These experimental results clearly demonstrate the effectiveness of the accurate bounding boxes and confirm that bounding boxes with noise could lead to a failure of object relation matrix construction, thus further decreasing the role of GCN1. Moreover, for the task of \textbf{PREDCLS}, the performance of R2-Net (w/o GCN1) and R2-Net is almost the same. As mentioned in the Ablation Study (i.e., Sec.~\ref{DEEPFE}), the object relation prediction is largely depending on the accuracy of the object labels instead of the object region features. Therefore, with ground truth object labels, using GCN1 to improve the region features could not considerably improve the performance of the object relation prediction process.

\subsection{Results and Analysis: Qualitative Results}

We show some qualitative examples in Fig.~\ref{Fig:qualitative} obtained by our R2-Net (w/o GCN1) on the \textbf{SGDET}. From the first column, we can see that our model not only predicts relationships in the ground truth but also predicts relationships not in the ground truth, such as \textit{(pot-1, has, handle-1)}, \textit{(boat-2, has, dog-1)},  \textit{(tie1, on, shirt-1)}, and \textit{(hair, on, head-1)}. From the second column, we can see that undetected bounding boxes are a major reason for leading to relation generation failures. For instance, \textit{railing-GT1} and \textit{logo-GT1} in the first row, \textit{tire-GT1} in the second row, \textit{paw-GT1} in the third row and \textit{leg-GT1} in the bottom, are not detected. Therefore, the relations correlated with them are not detected. 

We also show some affinity matrices ($\textbf{A}^e$ and $\textbf{A}^r$ mentioned in Eq.~\ref{equ.dnor3} and Eq.~\ref{equ.encodr2}) in Fig.~\ref{Fig:qualitative_adj}. From examples on the left side, both $\textbf{A}^e$ and $\textbf{A}^r$ can properly predict whether there is a relationship between instances. However, in the examples on the right side, the visual link prediction of $\textbf{A}^r$ is better than $\textbf{A}^e$. For instance, $\textbf{A}^r$ can provide correct links of the following object pairs: \textit{(plane-1, tail-1)}, \textit{(man-1, wheel-2)}, \textit{(wheel-1, skateboard-1)}, and \textit{(food-2, plate-1)}. In the right bottom example, \textit{food-2} (i.e., soup and bread) contains \textit{food-1} (i.e., bread), but the relationship category between \textit{food-2} and \textit{food-1} is not contained in the Visual Genome dataset. Therefore, in the affinity matrix $\textbf{A}^r$, the value of the object pair \textit{(food-1, food-2)} is small. The reason why the prediction of $\textbf{A}^r$ is better than $\textbf{A}^e$ includes two aspects. First, after the Object Label Refiner, the prediction of the object label is improved. Second, the residual connection allows the Object Relationship Generator to receive more robust features.

\section{Conclusion}
In this paper, we propose a novel model for parsing visual scene graphs, namely Relation Regularized Network (R2-Net). It predicts whether there is a relationship between two objects and generates an affinity matrix. GCNs over the affinity matrix aggregate the related object features to the target object features. In this way, the model can integrate the information of related objects to boost the performance of the label prediction.
BiLSTMs are used to extract the global contexts of objects. By encoding object features with global context and relational information, the relation regularized module can effectively refine the prior labels from Faster R-CNN and predict the relationships between objects. Compared with other methods, our model produces more accurate object labels and more robust relationship features, so our model outperforms the state-of-the-art methods on the scene graph generation task. Extensive experiments on the Visual Genome dataset demonstrate the effectiveness of our method.

\section{Acknowledgment}
This work is supported by the National Key Research and Development Program of China (Grant No. 2018AAA0102200), the National Natural Science Foundation of China (Grant No. 61772116, No. 61872064, No. 62020106008, No. 61871470), the Sichuan Science and Technology Program (Grant No. 2019JDTD0005), the Open Project of Zhejiang Lab (Grant No. 2019KD0AB05) and the Open Project of Key Laboratory of Artificial Intelligence, Ministry of Education (Grant No. AI2019005).

% or


\appendices

\ifCLASSOPTIONcaptionsoff
  \newpage
\fi



\bibliographystyle{IEEEtran}
\bibliography{ref}


\begin{IEEEbiography}
	[{\includegraphics[width=1in,height=1.25in]{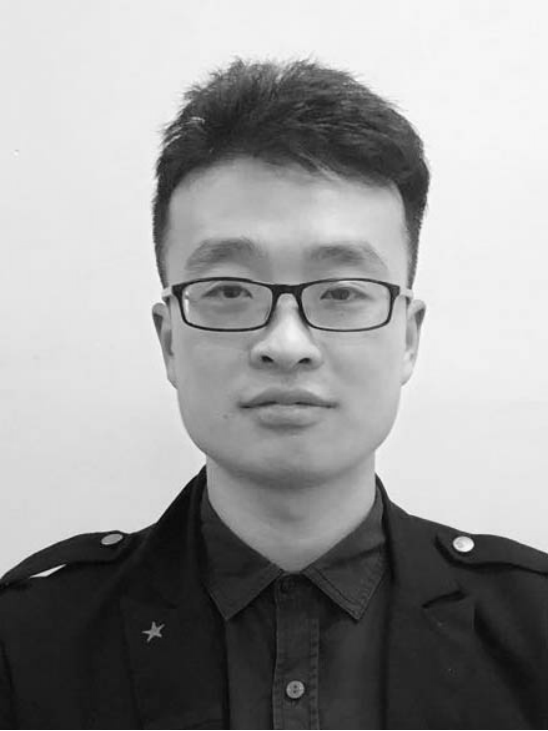}}]{Yuyu Guo}
	is a PH.D. student in the School of Computer Science and Engineering, University of Electronic Science and Technology of China. Currently, he is working on image understanding, image/video captioning and scene graph generation.
\end{IEEEbiography}

\begin{IEEEbiography}
	[{\includegraphics[width=1in,height=1.25in]{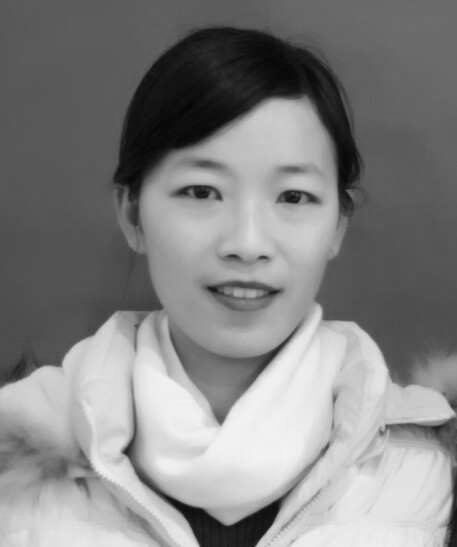}}]{Lianli Gao}
	(Member, IEEE)
	received the Ph.D. degree in information technology from The University of Queensland (UQ), Brisbane, QLD, Australia, in 2015. She is currently a Professor with the School of Computer Science and Engineering, University of Electronic Science and Technology of China (UESTC), Chengdu, China. She is focusing on integrating natural language for visual content understanding. Dr. Gao was the winner of the IEEE Trans. on Multimedia 2020 Prize Paper Award, the Best Student Paper Award in the Australian Database Conference, Australia, in 2017, the IEEE TCMC Rising Star Award in 2020, and the ALIBABA Academic Young Fellow.
\end{IEEEbiography}

\begin{IEEEbiography}
	[{\includegraphics[width=1in,height=1.25in]{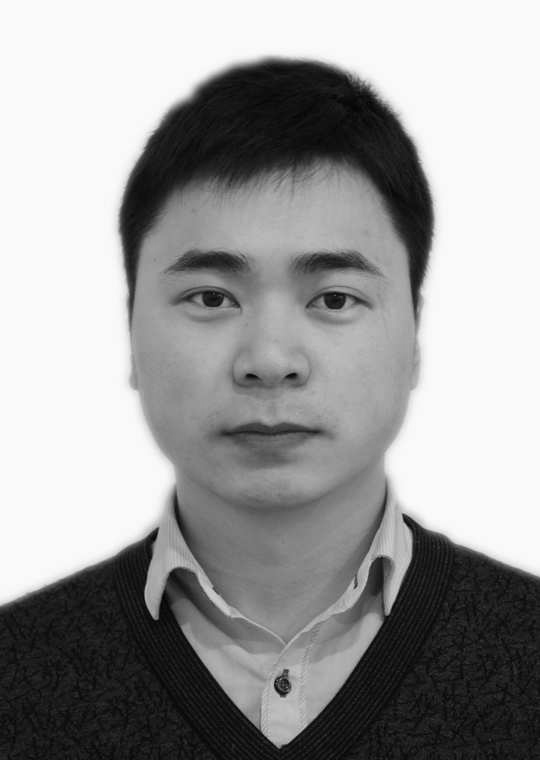}}]{Jingkuan Song} (Senior Member, IEEE) is currently a Professor with the University of Electronic Science and Technology of China (UESTC), Chengdu, China. His research interests include large-scale multimedia retrieval, image/video segmentation and image/video understanding using hashing, graph learning, and deep learning techniques. Dr. Song has been an AC/SPC/PC Member of IEEE Conference on Computer Vision and Pattern Recognition for the term 2018–2021, and so on. He was the winner of the Best Paper Award in International Conference on Pattern Recognition, Mexico, in 2016, the Best Student Paper Award in Australian Database Conference, Australia, in 2017, and the Best Paper Honorable Mention Award, Japan, in 2017.
\end{IEEEbiography}

\begin{IEEEbiography}
	[{\includegraphics[width=1in,height=1.25in]{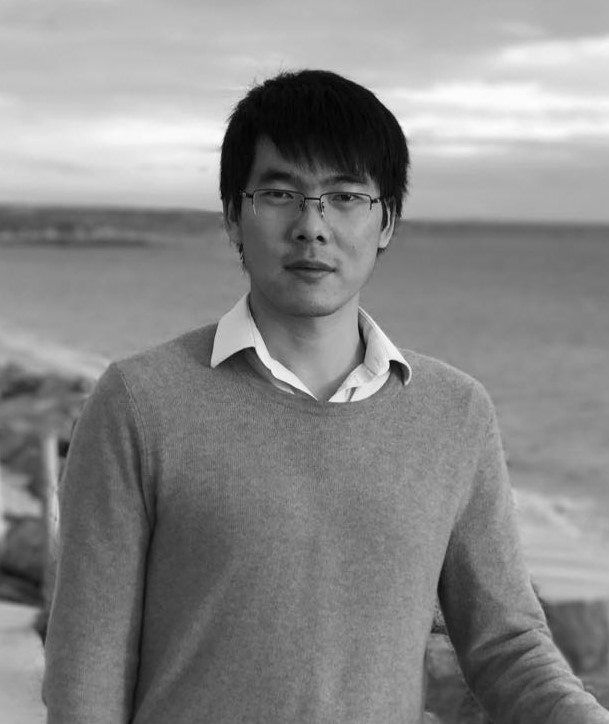}}]{Peng Wang} received the Ph.D. degree from the University of Queensland in 2017. He was a Research Fellow with the Australia Institute of Machine Learning (AIML), The University of Adelaide. He is currently a Lecturer with the School of Computing and Information Technology, University of Wollongong (UOW), Australia. His research interests include computer vision and deep learning, with focus on low-shot classification, long-tail classification, and visual reasoning. His research works have been published on main computer vision journals and conferences, such as TPAMI, IJCV, TIP, CVPR, and AAAI.
\end{IEEEbiography}

\begin{IEEEbiography}[{\includegraphics[width=1in,height=1.25in]{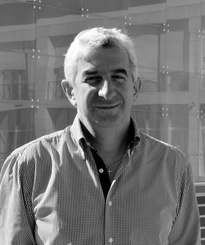}}]{Nicu Sebe}
	(Senior Member, IEEE) is Professor with University of Trento, Italy, leading the research in the areas of multimedia information retrieval and human behavior understanding. He was the General Co-Chair of the IEEE FG Conference 2008 and ACM Multimedia 2013, and the Program Chair of the International Conference on Image and Video Retrieval in 2007 and 2010, ACM Multimedia 2007 and 2011. He was the Program Chair of ICCV 2017 and ECCV 2016, and a General Chair of ACM ICMR 2017. He is a fellow of  IAPR.
\end{IEEEbiography}

\begin{IEEEbiography}
	[{\includegraphics[width=1in,height=1.25in]{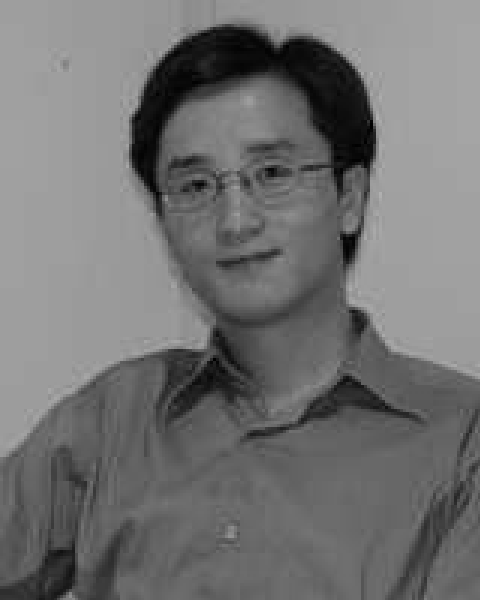}}]{Heng Tao Shen}
	(Fellow, ACM) received the B.Sc. (First Class Hons.) and Ph.D. degrees in computer science from the Department of Computer Science at National University of Singapore in 2000 and 2004 respectively. He is currently a Professor, the Dean of School of Computer Science and Engineering, the Executive Dean of Artificial Intelligence Research Institute, and the Director of Center for Future Media at University of Electronic Science and Technology of China. His current research interests include multimedia search, computer vision, artificial intelligence, and big data management. He has published over 250 peer-reviewed papers, and received 7 best paper awards from international conferences, including the Best Paper Award from ACM Multimedia 2017 and Best Paper Award-Honourable Mention from ACM SIGIR 2017. He has served as General Co-chair for ACM Multimedia 2021 and Program Committee Co-Chair for ACM Multimedia 2015, and is an Associate Editor of ACM Transactions of Data Science, IEEE Transactions on Image Processing, IEEE Transactions on Multimedia, and IEEE Transactions on Knowledge and Data Engineering.
\end{IEEEbiography}

\begin{IEEEbiographynophoto}
	{Xuelong Li} (M'02-SM'07-F'12) is a full professor with School of Computer Science and Center for OPTical IMagery Analysis and Learning (OPTIMAL), Northwestern Polytechnical University, Xi'an 710072, P.R. China.
\end{IEEEbiographynophoto}

%\vfill


\end{document}